\documentclass[10pt,conference]{IEEEtran}
\IEEEoverridecommandlockouts

\usepackage{cite}
\usepackage{amsmath,amssymb,amsfonts}
\usepackage{algorithmic}
\usepackage{graphicx}
\usepackage{textcomp}
\usepackage[english]{babel}
\usepackage{amsmath}
\usepackage{graphicx,epstopdf}
\usepackage[english]{babel}
\usepackage[ruled,vlined]{algorithm2e}
\usepackage{amsmath}
\usepackage{booktabs}
\usepackage{graphicx}
\usepackage{amsmath,amssymb}
\usepackage{latexsym}
\usepackage{eurosym}
\usepackage{units}
\usepackage{mathtools}
\usepackage{lscape} 
\usepackage{amssymb}
\usepackage{xcolor}
\usepackage{lipsum}
\usepackage{optidef}
\usepackage{multirow}
\usepackage{booktabs}

\def\BibTeX{{\rm B\kern-.05em{\sc i\kern-.025em b}\kern-.08em
    T\kern-.1667em\lower.7ex\hbox{E}\kern-.125emX}}
\usepackage{subcaption}
\usepackage{tabularx}
\newcommand{\revised}[1]{{\color{blue}#1}}

\usepackage{fancyhdr}
\usepackage{lipsum} 
\begin{document}

\title{
\
Intelligent Multi-UAV Navigation in ITNTNs: A Hierarchical LLM Approach
}

  


\author{%
  \IEEEauthorblockN{%
    Zijiang~Yan\IEEEauthorrefmark{1}, 
    Hao~Zhou\IEEEauthorrefmark{2}, 
    Wael~Jaafar\IEEEauthorrefmark{3}, 
    Jianhua~Pei\IEEEauthorrefmark{1}, 
    Ping~Wang\IEEEauthorrefmark{1}, 
    Halim~Yanikomeroglu\IEEEauthorrefmark{4}, 
    and Hina~Tabassum\IEEEauthorrefmark{1}%
  }%
  \fontsize{9}{10}\selectfont
  \IEEEauthorblockA{%
    \IEEEauthorrefmark{1}York University, Toronto, ON, Canada \quad 
    \IEEEauthorrefmark{2}Samsung Research America, Toronto, ON, Canada
  }
  \IEEEauthorblockA{%
    \IEEEauthorrefmark{3} ÉTS, University of Quebec, Montréal, QC, Canada \quad 
    \IEEEauthorrefmark{4} Carleton University, Ottawa, ON, Canada 
  }

\thanks{
Emails: 
 \{zijiang, pingw, hinat\}@yorku.ca,
 haozhou029@gmail.com,
 wael.jaafar@etsmtl.ca,
 jianhuapei98@gmail.com,
 halim@sce.carleton.ca.
}
}


\maketitle
\raggedbottom

\thispagestyle{fancy}
\begin{abstract}
The deployment of high-speed Uncrewed Aerial Vehicles (UAVs) in 3D aerial highways necessitates robust coordination of physical flight kinematics and multi-tier network handovers. While Deep Reinforcement Learning (DRL) offers rapid tactical control, it lacks the zero-shot strategic reasoning required to quickly adapt to dynamic Integrated Terrestrial and Non-Terrestrial Networks (ITNTNs). Conversely, Large Language Models (LLMs) excel at semantic reasoning but suffer from high inference latency, rendering them unsuitable for real-time aerodynamic control. To bridge this gap, we propose a novel Hierarchical LLM-driven control framework. A massive cloud-based LLM deployed on a High-Altitude Platform Station (HAPS) manages slow-timescale global load balancing, while lightweight edge-LLMs on individual UAVs translate local observations into tactical sub-goals. These sub-goals guide a fast-timescale physical DRL controller to execute collision-free, handover-aware trajectories. Simulation results demonstrate that our agentic architecture significantly reduces collision rates and improves aggregate system throughput compared to existing baselines.
\end{abstract}


\section{Introduction}
\label{sec:introduction}

The rapid proliferation of Uncrewed Aerial Vehicles (UAVs) is driving the conceptualization of 3D aerial highways, dedicated corridors designed to support dense, high-speed autonomous logistics \cite{cherif20213d}. Operating safely in these environments requires uninterrupted Command and Control (C2) and telemetry links \cite{yan2023multi}. To satisfy these stringent connectivity requirements, 6G architectures are shifting toward Integrated Terrestrial and Non-Terrestrial Networks (ITNTNs) \cite{JaafarHAPSITS}, leveraging High-Altitude Platform Stations (HAPS) to complement the terrestrial base stations' (TBSs) coverage. However, as UAVs traverse these overlapping coverage tiers at high velocities, they inevitably trigger frequent network handovers (HOs) \cite{yan2022reinforcement}, necessitating the joint optimization of physical flight kinematics and communication reliability.

Although Deep Reinforcement Learning (DRL) has been widely adopted for UAV trajectory planning and cell association \cite{Cherif2024}, it inherently relies on trial-and-error exploration \cite{yan2023multi}. In multi-agent aerial corridors, such exploration often leads to localized and greedy optimization, which can result in the systemic depletion of shared resources. Uncoordinated agents may inadvertently overload specific TBSs or saturate the limited bandwidth of the HAPS, ultimately compromising both the aggregate system capacity and flight safety \cite{jaafar2020dynamics,yan2025cvar}.


Recently, the integration of Generative AI, especially Large Language Models (LLMs), has emerged as a transformative approach for autonomous vehicle control \cite{dong2026aerial}. LLMs possess unparalleled zero-shot reasoning, enabling complex mission adaptation and global strategic planning \cite{yan2025hierarchical}. Yet, standard LLM inference suffers from significant latency bottlenecks. Relying purely on LLMs is fundamentally incompatible with the fast-timescale execution required for sub-millisecond RF channel handovers and high-frequency rotor control \cite{chen2021bdfl,yan2025hybrid}. Bridging the gap between slow-timescale semantic reasoning and fast-timescale physical execution is an open challenge.

To overcome these limitations, this paper proposes a novel \textbf{Dual-Timescale Hierarchical LLM architecture} that synergizes the cognitive reasoning of foundation models with the rapid execution of DRL. 
Indeed, given the potential of Agentic AI and the latency challenges inherent in dynamic aerial networks, we propose here a novel Cloud-Edge framework that jointly optimizes UAV mobility and handover management. In particular, we first formulate the problem as a Hierarchical Multi-Objective Partially Observable Markov Decision Process (H-MO-POMDP), which strictly accounts for high-fidelity 3D rigid-body kinematics and realistic ITNTN capacity constraints. To address this complex hierarchy, we introduce a dual-timescale architecture powered by LLMs. Specifically, a strategic meta-controller (\texttt{Qwen3.5-122B} \cite{yang2025qwen3}) deployed on the HAPS operates at a slow timescale to dictate global load-balancing policies. Moreover, unlike previous studies that relied solely on trial-and-error DRL, we integrate here lightweight edge-agents (\texttt{Qwen3.5-9B} \cite{yang2025qwen3}) directly on individual UAVs. The latter generate cognitive reflections to dynamically tune the reward functions of a fast-timescale Deep Double Q-Network (DDQN), successfully bridging semantic reasoning with real-time physical execution. Finally, extensive simulations in a high-fidelity 3D physics environment demonstrate that the proposed LLM-guided framework mitigates severe collision penalties and maximizes system throughput, thus outperforming baselines.
%

{The remainder of this paper is organized as follows. Section \ref{sec:system_model} describes the ITNTN system model, including the high-fidelity 3D UAV kinematics and the multi-tier communication architecture. Section \ref{sec:problem_formulation} formulates the joint optimization problem as a H-MO-POMDP. The proposed dual-timescale hierarchical LLM framework and its cognitive reasoning modules are introduced in Section \ref{sec:llm_framework}. Section \ref{sec:numerical_results} presents the numerical results and performance evaluations. Finally, Section \ref{sec:conclusion} concludes the paper.}

\section{System Model}
\label{sec:system_model}
\begin{figure}[t]
    
    \centering
    \includegraphics[trim={1.2cm 0 0.7cm 0},clip,width=1\linewidth]{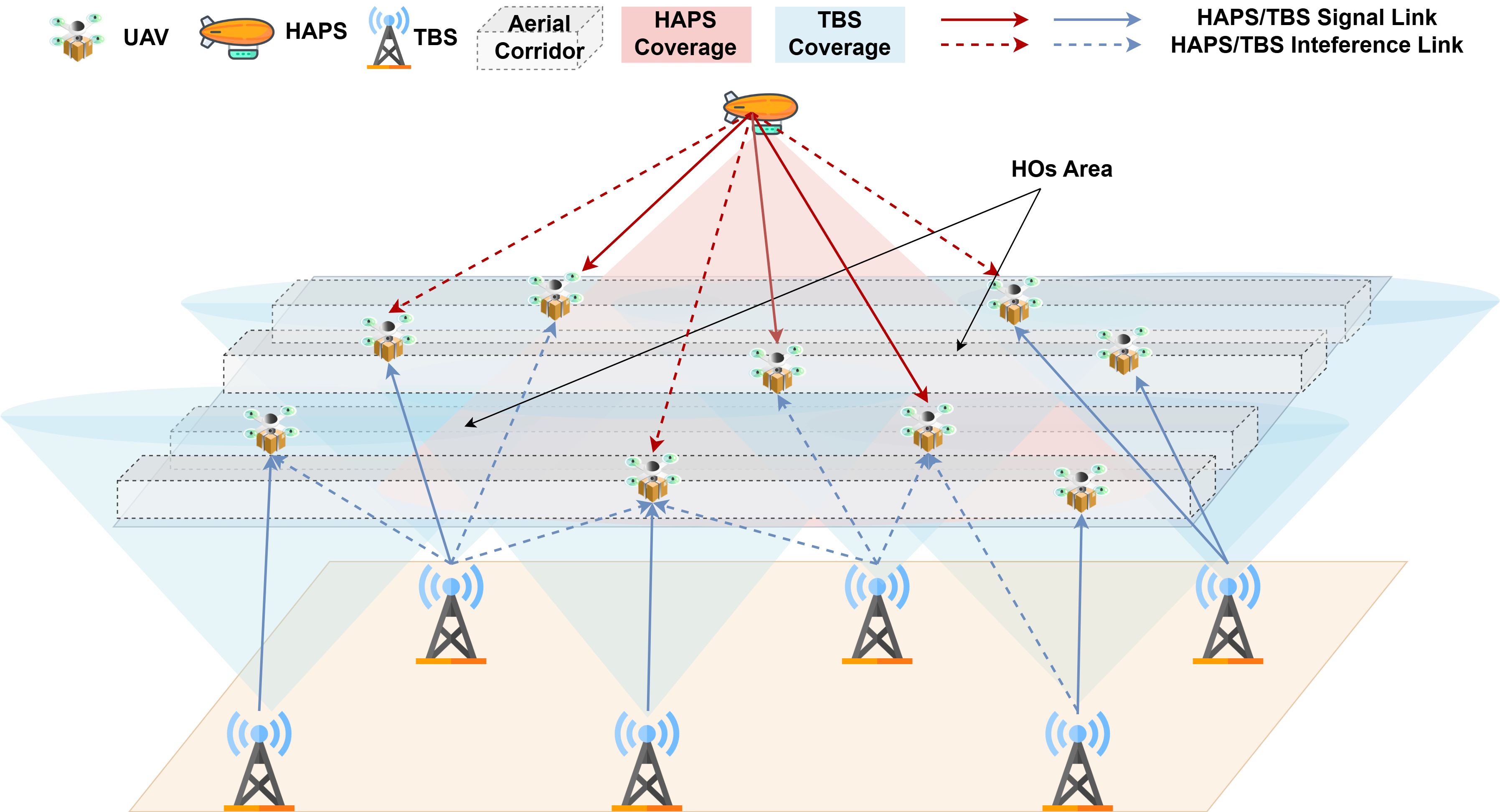}
    \caption{3D aerial network model. 
    \label{fig:Architecture}
    }

\end{figure}
\subsection{3D Aerial Highway and ITNTN Architecture}
As depicted in Fig.~\ref{fig:Architecture}, we consider a dynamic 3D aerial highway where a set of $M$ UAVs, denoted by $\mathcal{M}= \{1, 2, \dots ,M\}$, navigate a shared airspace. Each UAV $m \in \mathcal{M}$ is assigned an independent transit mission to a unique 3D target waypoint, $\mathbf{x}_{\text{target}}^m$, and must continuously adjust its altitude and heading to bypass obstacles and avoid collisions. 

To support mission-critical C2 and telemetry, the airspace is covered by an ITNTN. The network comprises a set of $B$ TBSs, denoted as $\mathcal{B} = \{1, 2, \dots ,B\}$, and a one HAPS, denoted by $H$. A UAV $m$ can connect to any serving node $c \in \mathcal{B} \cup \{H\}$, but high-speed traversal triggers handovers.

\subsection{Communication Model}
To evaluate the reliability of the C2 links, we formulate the signal-to-interference-plus-noise ratio (SINR)\cite{cherif2022cellular}. Let $G_t^{m,c}$ denote the linear channel power gain between UAV $m$ and serving node $c$ at time $t$, encapsulating both large-scale path loss and small-scale Rician fading. The received downlink SINR at UAV $m$ is given by
\begin{equation}
    \text{SINR}_t^{m,c} = \frac{P_c G_t^{m,c}}{N_0 B_c + I_t},
\end{equation}
where $P_c$ is the transmit power of node $c$, $N_0$ is the noise power spectral density, and $B_c$ is the allocated bandwidth. The term $I_t = \sum_{c' \neq c} P_{c'} G_t^{m,c'}$ represents the aggregate interference from other active transmitting nodes in the tier. 

The achievable data rate for UAV $m$ is calculated as
\begin{equation}
    R_t^{m,c} = B_c \log_2\left(1 + \text{SINR}_t^{m,c}\right).
\end{equation}
To penalize the latency and packet drops associated with frequent network switching, we define the handover-aware weighted data rate as
\begin{equation}
    \text{WR}_t^{m,c} = R_t^{m,c} - \gamma \mathbb{I}_{\text{HO},t}^m,
\end{equation}
where $\mathbb{I}_{\text{HO},t}^m \in \{0, 1\}$ is a binary indicator that equals $1$ if a handover occurs at time $t$, and $\gamma$ is the handover penalty coefficient. Also, the HAPS is constrained by a maximum aggregate capacity $C_{\max}^{\mathrm{HAPS}}$ and any node $c$ has a users association capacity $Q_c$, $\forall c\in \mathcal{B} \cup \{H\}$.

\subsection{Kinematic Model}
We adopt a high-fidelity 3D rigid-body kinematics model, where the physical state of UAV $m$ at time $t$ is defined by its 3D position $\mathbf{x}_t^m \in \mathbb{R}^3$ and linear velocity $\mathbf{v}_t^m \in \mathbb{R}^3$. 
The system evolves at a fast discrete timescale $\Delta t$ (e.g., =0.05 s), and the continuous kinematic state updates are governed by \cite{panerati2021learning}
\begin{equation}
\label{eq:x_t}
    \mathbf{x}_{t+1}^m = \mathbf{x}_t^m + \mathbf{v}_t^m \Delta t,
\end{equation}
\begin{equation}
\label{eq:v_t}
    \mathbf{v}_{t+1}^m = \mathbf{v}_t^m + \left( \frac{\mathbf{F}_{T,m}}{M_u} - \mathbf{g} - \mathbf{d}_m \right) \Delta t,
\end{equation}
where $M_u$ is the UAV mass, $\mathbf{g}$ is the gravitational acceleration vector, and $\mathbf{d}_m$ is the aerodynamic drag. The collective thrust vector $\mathbf{F}_{T,m}$ is derived from the real-time rotational speeds (RPM) of the four individual rotors. 
To ensure flight safety, a severe collision penalty is incurred if the 3D Euclidean distance between any two UAVs falls below the physical safety threshold: $\| \mathbf{x}_t^m - \mathbf{x}_t^j \|_2 < d_{\text{safe}}$ for any $m \neq j$.

\section{Problem Formulation}
\label{sec:problem_formulation}

To capture the coupled dynamics of flight kinematics and network handovers across different operational timescales, we formulate the system as a  H-MO-POMDP consisting of a slow-timescale global meta-controller (HAPS) and fast-timescale tactical edge-agents (UAVs).

\subsection{State and Observation Spaces}
Due to the limited sensing range of individual UAVs, the global environmental state $\mathcal{S}_t$ is partially observable. At the fast tactical timescale $t$, each UAV $m$ receives a local observation $\mathbf{o}_t^m \in \Omega$, defined as
\begin{equation}
    \mathbf{o}_t^m = \big[ \mathbf{x}_t^m, \mathbf{v}_t^m, \mathbf{x}_{\text{target}}^m, \{ \mathbf{x}_t^j, \mathbf{v}_t^j \}_{j \in \mathcal{N}_t^m}, c_t^m, \text{WR}_t^{m,c} \big],
\end{equation}
where $\mathcal{N}_t^m$ is the set of neighboring UAVs within sensing range, $c_t^m$ is the current serving node, and $\text{WR}_t^{m,c}$ is the handover-aware weighted data rate.
Conversely, at the slow strategic timescale $T_{\mathrm{HAPS}}$, the HAPS receives a global meta-observation $\mathbf{o}_T^{\mathrm{HAPS}}$ comprising the aggregate traffic load on each TBS, the available HAPS backhaul capacity, and the spatial distribution of the UAV swarm.

\subsection{Hybrid Action Spaces}
The action space encompasses both continuous physical controls and discrete network decisions as follows.

\subsubsection{HAPS Strategic Action} At each macro-step $T_{\mathrm{HAPS}}$, the HAPS agent outputs a discrete meta-action $\mathbf{a}_T^{\mathrm{HAPS}} \in \{\text{Offload}, \text{Recall}, \text{Idle}\}$. These directives are broadcast to specific high-density congestion zones to enforce load-balancing before node capacities are breached.

\subsubsection{UAV Tactical Action} At each micro-step $t$, UAV $m$ executes a joint hybrid action $\mathbf{a}_t^m = [\mathbf{a}_{\text{mot},t}^m, \mathbf{a}_{\text{tele},t}^m]^\top$. The continuous motion action $\mathbf{a}_{\text{mot},t}^m \in [-1, 1]^4$ regulates the RPM of the four individual rotors for physical 3D navigation. Concurrently, the discrete telecommunication action $\mathbf{a}_{\text{tele},t}^m \in \mathcal{B} \cup \{H\}$ selects the target serving node, triggering a handover if $a_{\text{tele},t}^m \neq c_t^m$.

\subsection{Multi-Objective Reward Functions}
To simultaneously optimize mobility and connectivity, we define scalarized reward structures for edge and HAPS levels.

\subsubsection{UAV Edge Reward} The objective of UAV $m$ is to safely navigate to its target while maintaining a robust C2 link. The reward is formulated as
\begin{equation}
    R_{t}^m = \alpha_1 R_{\text{tran},t}^m + \alpha_2 \text{WR}_t^{m,c} + \alpha_3 \rho_{\text{crash}},
\end{equation}
where $R_{\text{tran},t}^m = \exp(-\| \mathbf{x}^m_t - \mathbf{x}_{\text{target}}^m \|_2) - \lambda \| \mathbf{a}_{\text{mot}, t}^m \|_2^2$ rewards forward progression while penalizing aggressive, energy-consuming rotor maneuvers. The term $\text{WR}_t^{m,c}$ maximizes the data rate while penalizing handovers. Finally, $\rho_{\text{crash}}$ applies a severe negative penalty if the safety distance $d_{\text{safe}}$ is violated. The weights $\alpha_{\{1,2,3\}}$ balance the rewrards/penalties.

\subsubsection{HAPS Reward} The objective of the HAPS meta-controller is to maximize aggregate system throughput while enforcing node capacity limits. Its reward is given by
\begin{equation}
    R_{T}^{\mathrm{HAPS}} = \eta_1 \sum_{m \in \mathcal{M}} R_t^{m,c} - \eta_2 \sum_{c \in \mathcal{B} \cup \{H\}} \max(0, n_{c,t} - Q_c),
\end{equation}
where $n_{c,t}$ is the number of UAVs currently associated with node $c$. The second term heavily penalizes any policy that results in network congestion (i.e., exceeding the capacity $Q_c$), balanced by the weights $\eta_{\{1,2\}}$.
{
To capture the coupled dynamics of 3D trajectory control and handover-aware cell association, we define the joint optimization problem $\mathcal{P}$ below. The objective is to maximize the aggregate system utility while adhering to physical kinematics and network capacity constraints.

\begin{equation}
\resizebox{1\linewidth}{!}{$
\begin{aligned}
\mathcal{P}: \quad
\max_{\mathbf{A}, \boldsymbol{\alpha}} \quad
& \mathbb{E} \left[
\sum_{t=0}^{T} \gamma^t
\sum_{m \in \mathcal{M}}
\mathcal{R}_t^m(\mathbf{a}_t^m, \boldsymbol{\alpha}_t^m)
\right] \\
\text{s.t.} \quad
& \mathrm{C1}:~
\mathbf{x}_{t+1}^m =
f(\mathbf{x}_t^m,\mathbf{v}_t^m,\mathbf{a}_{\mathrm{mot},t}^m),
&& \forall m,t, \\
& \mathrm{C2}:~
\|\mathbf{x}_t^m-\mathbf{x}_t^j\|_2 \ge d_{\mathrm{safe}},
&& \forall m \ne j, \\
& \mathrm{C3}:~
\sum_{c \in \mathcal{B}\cup\{H\}} u_t^{m,c}=1,
&& \forall m,t, \\
& \mathrm{C4}:~
\sum_{m \in \mathcal{M}} u_t^{m,c}\le Q_c,
&& \forall c \in \mathcal{B}\cup\{H\}, \\
& \mathrm{C5}:~
\boldsymbol{\alpha}_t^m \in \Delta^3,\quad
\sum_{i=1}^{3}\alpha_{t,i}^m=1.
\end{aligned}
$}
\end{equation}

%
{In ($\mathcal{P}$), $\mathbf{A} \triangleq \{\mathbf{a}_t^m\}$ and $\boldsymbol{\alpha} \triangleq \{\boldsymbol{\alpha}_t^m\}$ ($\forall m \in \mathcal{M}, \forall t$) denote the joint hybrid action policy and the dynamically tuned multi-objective reward weights across all UAVs over the operational horizon, respectively.}
 $\mathbf{a}_t^m = [\mathbf{a}_{\text{mot},t}^m, a_{\text{tele},t}^m]^\top$ represents the hybrid action vector. \textbf{C1} ensures the trajectory updates follow the 3D rigid-body physics model defined in (\ref{eq:x_t})-(\ref{eq:v_t}).
\textbf{C2} maintains the physical safety separation between any two UAVs.
\textbf{C3} ensures that each UAV $m$ is associated with exactly one serving node (TBS or HAPS) at any time $t$.
\textbf{C4} restricts the number of associated users per node to its capacity $Q_c$ to prevent network saturation.
\textbf{C5} defines the simplex $\Delta^3$ for the cognitive reward weights $\alpha_{\{1,2,3\}}$ tuned by the edge-LLM.
%

{Problem $\mathcal{P}$ is a highly non-convex mixed-integer non-linear programming problem coupling flight kinematics and aerial communications. To enable real-time control, we decompose this optimization via a two-tier Hierarchical Multi-Objective POMDP framework, delegating specific constraints to distinct LLM cognitive tiers, as described in the next section.}
}

\section{Proposed Hierarchical LLM Framework}
\label{sec:llm_framework}


\begin{figure*}[http]
    \centering
    \includegraphics[width=0.5\linewidth]{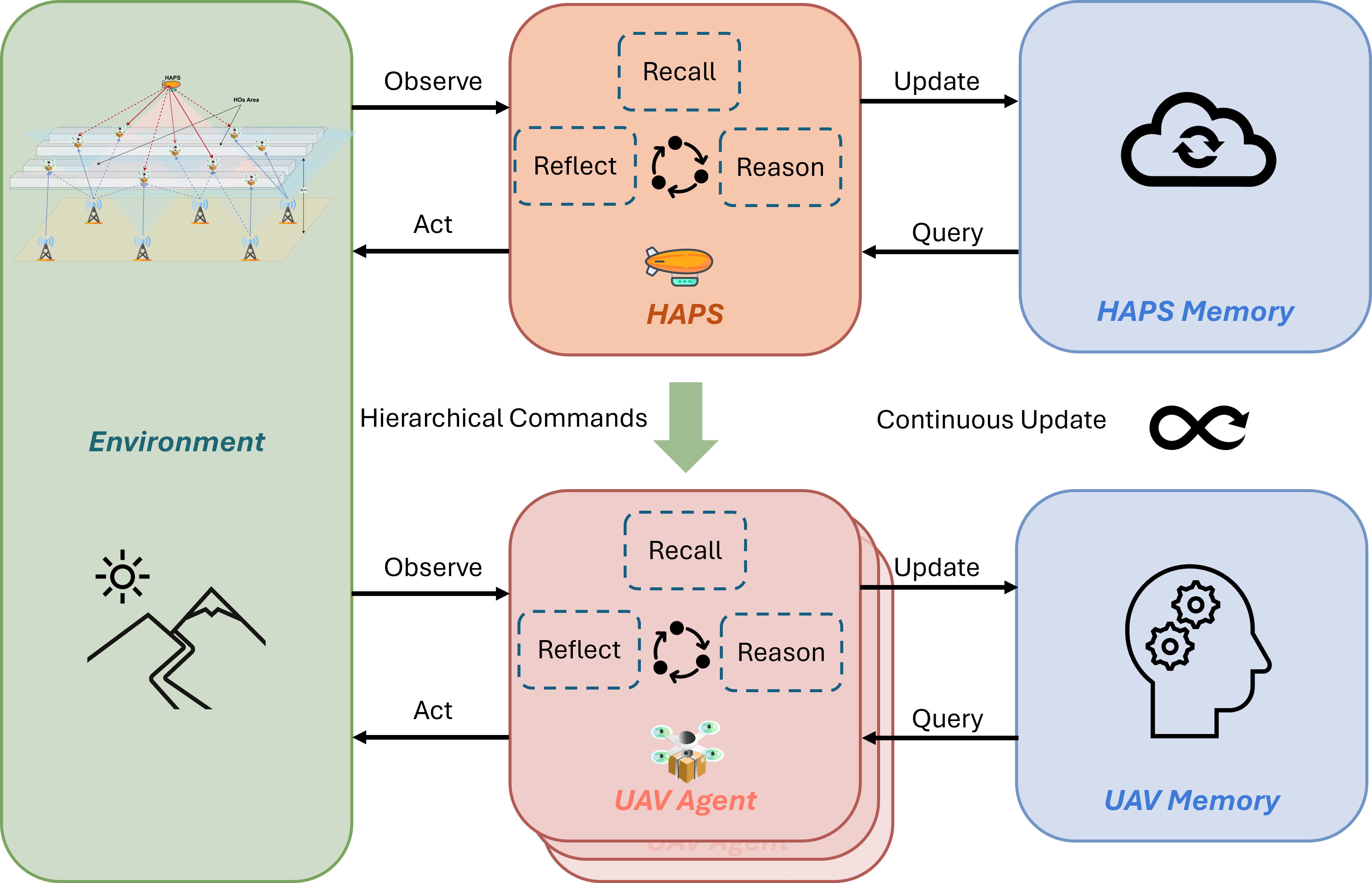}
    \caption{Hierarchical control for UAVs in an ITNTN.}
    \label{fig:haps_hierarchical_framework}
    \vspace{-9pt}
\end{figure*}

To bridge the latency gap between semantic reasoning and real-time aerodynamic control, we propose a Cloud-Edge Agentic architecture, as illustrated in Fig.~\ref{fig:haps_hierarchical_framework}. The framework separates decision-making into three distinct timescales: 1) slow strategic meta-timescale ($T_{\mathrm{HAPS}}$), 2) intermediate tactical reasoning timescale ($T_{\mathrm{LLM}}$), and 3) fast physical execution timescale ($T_{\mathrm{fast}}$).

\subsection{HAPS Cloud Meta-Controller }
At the highest level of the hierarchy, the HAPS acts as a global orchestrator. Since global load balancing requires complex spatial reasoning across the entire ITNTN, we deploy a large foundation model, i.e., \texttt{Qwen3.5-122B}, hosted in the cloud computing cluster of HAPS. 
Operating at $T_{\mathrm{HAPS}} = 5.0$~s, the HAPS meta-controller evaluates the global observation $\mathbf{o}_T^{\mathrm{HAPS}}$. In particular, if the number of users associated with a specific TBS $n_{c,t}$ is close to $Q_c$, the LLM generates zero-shot strategic directives ($\mathbf{a}_T^{\mathrm{HAPS}}$), forcing specific UAV edge-agents to offload their C2 links to neighboring TBSs or the HAPS.

\subsection{UAV Edge-Agent }
Each UAV is equipped with a lightweight edge-LLM (i.e., \texttt{Qwen3.5-9B}) that operates at $T_{\mathrm{LLM}} = 1.0$~s. 
The cognitive edge-agent translates the localized observation ($\mathbf{o}_t^m$) into a structured textual prompt. Rather than outputting direct motor controls, the edge-LLM performs \textit{cognitive reflection}, i.e., if the UAV experiences severe handover penalties or is close to a collision threshold ($d_{\text{safe}}$), the LLM dynamically adjusts the weights $\alpha_{\{1,2,3\}}$ of the multi-objective reward vector. By doing so, this semantic reasoning guides the underlying DRL, shifting its priority between aggressive forward transit and conservative safety.

\subsection{Fast-Timescale DRL Execution}
At the lowest level, a DDQN operates at the physics timescale $T_{\mathrm{fast}} = 0.05$~s (20 Hz). The DDQN receives the dynamic reward weights generated by the edge-LLM and executes the continuous motor commands ($\mathbf{a}_{\text{mot},t}^m$) and discrete handover actions ($\mathbf{a}_{\text{tele},t}^m$). By offloading semantic reasoning to the LLM, the DDQN acts purely as a high-speed, reflexive physical controller. 

The complete execution flow is summarized in Algorithm~\ref{algo:hierarchical}.

\begin{algorithm}[!htbp]
\caption{Dual-Timescale Hierarchical LLM Control}
\label{algo:hierarchical}
\small
\textbf{Initialize:} Environment state $\mathcal{S}_0$, HAPS LLM (\texttt{Qwen-122B}), UAV LLMs (\texttt{Qwen-9B}), and DDQN networks. \\
\For{each physics micro-step $t = 1, 2, \dots, T_{\max}$}{
    \If{$t \pmod{T_{\mathrm{HAPS}}} == 0$}{
        Extract global meta-observation $\mathbf{o}_T^{\mathrm{HAPS}}$\;
        HAPS LLM generates strategic meta-action $\mathbf{a}_T^{\mathrm{HAPS}}$ to balance node quotas $Q_c$\;
    }
    \For{each UAV $m \in \mathcal{M}$}{
        Extract local numerical observation $\mathbf{o}_t^m$\;
        \If{$t \pmod{T_{\mathrm{LLM}}} == 0$}{
            Format $\mathbf{o}_t^m$ into dynamic text prompt $\mathbf{p}_t^{\mathrm{dyn}}$\;
            Edge LLM evaluates safety threshold $\rho_{\text{thresh}}$\;
            Edge LLM outputs cognitive reflection to dynamically tune reward weights $\alpha_{\{1,2,3\}}$\;
        }
        DDQN receives tuned rewards and selects hybrid action $\mathbf{a}_t^m = [\mathbf{a}_{\text{mot},t}^m, \mathbf{a}_{\text{tele},t}^m]^\top$\;
        Execute $\mathbf{a}_t^m$, update kinematics at $T_{\mathrm{fast}}$, and store transition in memory buffer $\mathcal{D}^m$\;
    }
    Sample mini-batch from $\mathcal{D}^m$ to train DDQN\;
}
\end{algorithm}

\subsection{State Discretization and Prompt Engineering}
To bridge the gap between numerical kinematics and the semantic reasoning of the \texttt{Qwen3.5-9B} edge-LLM, we leverage a structured prompt engineering methodology. Directly feeding high-precision floating-point arrays (e.g., $\mathbf{v}_t^m = [12.43, -3.21, 0.5]$) often degrades LLM reasoning and inflates inference latency. Therefore, we discretize the continuous observation $\mathbf{o}_t^m$ into semantic language descriptors, denoted as $\mathcal{L}(\mathbf{o}_t^m)$.
Specifically, the 3D relative distance to neighboring UAVs ($\|\mathbf{x}_t^m - \mathbf{x}_t^j\|_2$, $\forall j \in \mathcal{M} \backslash \{ m \}$) is categorized into predefined semantic zones: \textit{Safe} ($d > 3 d_{\text{safe}}$), \textit{Warning} ($d_{\text{safe}} \le d \le 3 d_{\text{safe}}$), and \textit{Critical} ($d < d_{\text{safe}}$). 
{Similarly, the handover-aware network state $\text{WR}_t^{m,c}$ is mapped to \textit{Stable} (when $\text{WR}_t^{m,c} \geq 20$ Mbps, indicating an un-congested link), \textit{Degraded} (when $0 \le \text{WR}_t^{m,c} < 20$ Mbps, indicating capacity saturation), or \textit{Hand-over} (when $\text{WR}_t^{m,c} < 0$, triggered by the massive $\gamma$ penalty during network switching).}



To enhance the zero-shot generalization of the LLM, we utilize distance-based few-shot example selection. Rather than using static prompts, the edge-agent retrieves a set of $K$ historical states from its local memory buffer $\mathcal{D}^m$ that exhibit the shortest Euclidean distance to the current kinematic state. The successful reward-weight configurations from these retrieved states are appended as in-context learning examples. 

The final synthesized prompt strictly follows a modular template: \texttt{[Role]}, \texttt{[Current State]}, \texttt{[Historical Examples]}, and \texttt{[Task]}. An illustrative example of the generated prompt is provided below:

\noindent\fbox{%
    \parbox{0.95\columnwidth}{
        \scriptsize \raggedright
        \textbf{[Role]:} You are an autonomous UAV edge-agent navigating a 3D corridor. Your task is to dynamically tune the multi-objective DRL reward weights $(\alpha_1, \alpha_2, \alpha_3)$ for Transit Efficiency, Network Reliability, and Safety.\\
        \textbf{[Current State]:} Distance to target is \textit{Approaching}. Neighbor UAV-3 is in the \textit{Warning} zone (7.2m). Network link to TBS-2 is \textit{Stable}.\\
        \textbf{[Historical Example]:} When Neighbor was \textit{Warning} and Link was \textit{Stable}, optimal weights were $(0.2, 0.1, 0.7)$ to heavily prioritize collision avoidance over forward transit.\\
        \textbf{[Task]:} Based on the current state, output the updated scalarization weights in the exact format [w1, w2, w3]. Do not output conversational text or explanations.
    }
}
\vspace{2mm}

\section{Numerical Results}
\label{sec:numerical_results}

\subsection{Simulation Setup}
We evaluate the proposed Cloud-Edge Agentic framework using a high-fidelity 3D multi-rotor physics simulator based on \textit{gym-pybullet-drones} \cite{panerati2021learning}. The ITNTN environment consists of 4 TBSs and 1 HAPS serving a number of UAVs $M \in \{10, 20, 30\}$. The Qwen-based LLMs interact with the physical simulation via structured API calls. The detailed simulation parameters are summarized in Table~\ref{tab:sim_parameters}. 

A key challenge in applying generative AI to high-speed UAV networks is LLM inference latency, which is incompatible with the fast reaction time required for aerodynamic stabilization. To address this issue, we adopted a two-level HAPS--UAV architecture with separated cognitive and control timescales.
At the UAV edge, we consider a 4-bit quantized \texttt{Qwen3.5-9B} deployed on an edge-AI platform. With an average generation speed of about $25$ tokens/s, a structured prompt and a concise output yield an estimated edge latency of $t_{\text{edge}}=0.6$~s. At the HAPS layer, the larger \texttt{Qwen3.5-122B} meta-controller runs on a high-performance GPU cluster, with total delay modeled as $t_{\text{cloud}} = t_{\text{inf}} + t_{\text{tx}} = 2.55$~s.

These values mathematically motivate our proposed timescale hierarchy. The fast aerodynamic loop runs at $T_{\mathrm{fast}} = 0.05$~s, making direct LLM inference in the control path physically impractical. We therefore set the UAV-level cognitive reflection interval to $T_{\mathrm{LLM}} = 1.0$~s and the HAPS-level coordination interval to $T_{\mathrm{HAPS}} = 5.0$~s, strictly satisfying the hardware bounds $T_{\mathrm{LLM}} > t_{\text{edge}}$ and $T_{\mathrm{HAPS}} \gg t_{\text{cloud}}$. During one edge-LLM inference interval, the DDQN controller can still execute 20 fast, uninterrupted control steps using the latest validated reward configuration. This asynchronous decoupling avoids computation-induced crashes while enabling robust LLM strategic reasoning.

In our simulations, we compare the proposed Hierarchical LLM-DDQN framework against the {``DDQN''} baseline \cite{yan2022reinforcement, zhang2024machine}, which relies on trial-and-error exploration without LLM cognitive reflection or HAPS load-balancing, and the {``Envelope MORL''} method proposed in \cite{yan2024generalized}. 

\vspace{0.02in}
\begin{table}[http]
    \caption{Simulation Parameters}
    \centering
    \scriptsize
\begin{tabularx}{\columnwidth}{@{}l X@{}}
        \toprule
        \textbf{Parameter} & \textbf{Value} \\
        \midrule
        \multicolumn{2}{@{}l}{\textbf{Physics \& Airspace Parameters}} \\
        \midrule
        Simulation airspace volume & $1000 \times 1000 \times 300$ m$^3$ \\
        Number of UAVs ($M$)       & $\{5, 10, 15, 20, 25, 30\}$ \\
        UAV Mass ($M_u$)           & $1.5$ kg \\
        Physical safety separation ($d_{\text{safe}}$) & $5.0$ m \\
        Gravity ($\mathbf{g}$)     & $9.81$ m/s$^2$ \\
        Control frequency ($1/T_{\text{fast}}$) & $20$ Hz ($\Delta t = 0.05$ s) \\
        \midrule
        \multicolumn{2}{@{}l}{\textbf{ITNTN Communication Parameters}} \\
        \midrule
        Number of TBSs ($B$)       & $4$ \\
        HAPS altitude              & $20$ km \\
        Carrier frequencies        & $2.0$ GHz (HAPS), $2.1$ GHz (TBS) \\
        Node bandwidth ($B_c$)     & $20$ MHz \\
        HAPS total capacity ($C^{\text{HAPS}}_{\max}$) & $100$ Mbps \\
        Node transmit power ($P_c$)& $40$ dBm ($10,000$ mW) \\
        TBS peak antenna gain ($G_{\max}$) & $8$ dBi \\
        Node capacity quota ($Q_c$)& $5$ concurrent users \\
        Path loss exponents        & $\eta_{\text{LoS}} = 2.0$, $\eta_{\text{NLoS}} = 3.5$ \\
        Rician LoS factor          & $15$ dB \\
        Noise power spectral density ($N_0$) & $-174$ dBm/Hz \\
        \midrule
        \multicolumn{2}{@{}l}{\textbf{Algorithm \& Cognitive Parameters}} \\
        \midrule
        HAPS Meta-Controller LLM   & \texttt{Qwen3.5-122B} \\
        UAV Edge-Agent LLM         & \texttt{Qwen3.5-9B} \\
        HAPS strategic timescale ($T_{\mathrm{HAPS}}$) & $5.0$ s \\
        UAV LLM reasoning interval ($T_{\text{LLM}}$) & $1.0$ s \\
        Edge inference latency ($t_{\text{edge}}$) & $0.6$ s \\
        Cloud inference latency ($t_{\text{cloud}}$) & $2.55$ s \\
        Reflection trigger threshold ($\rho_{\text{thresh}}$) & $-10.0$ \\
        Collision crash penalty ($\rho_{\text{crash}}$) & $-100.0$ \\
        HAPS reward weights ($\eta_{\{1,2\}}$) & $\{1.0, 50.0\}$ \\
        UAV reward weights ($\alpha_{\{1,2,3\}}$) & $\{1.0, 0.1, 0.2\}$ \\
        Handover penalty coeff. ($\gamma$) & $5.0$ \\
        Energy penalty coeff. ($\lambda$)  & $0.1$ \\
        Memory retrieval size ($K$)& $5$ experiences \\
        DDQN learning rate         & $1 \times 10^{-4}$ \\
        Discount factor ($\gamma_d$) & $0.99$ \\
        Target update frequency    & $500$ steps \\
        \bottomrule
    \end{tabularx}
    \label{tab:sim_parameters}
\end{table}

\begin{figure*}[t]
    \centering

    \begin{subfigure}[t]{0.48\linewidth}
        \centering
        \includegraphics[width=\linewidth]{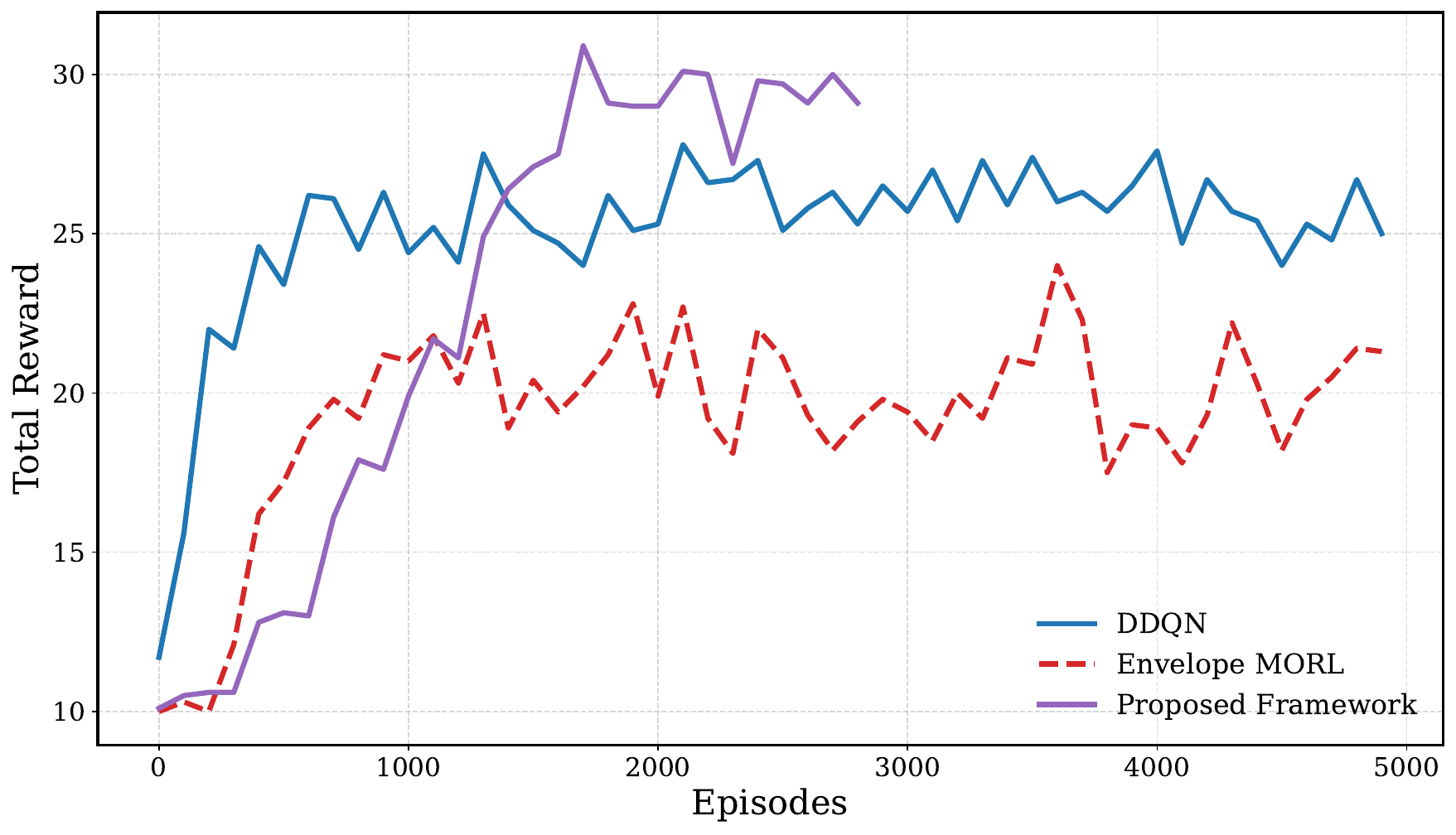}
        \caption{Transportation Reward ($R_{\text{tran},t}^m$)}
        \label{fig:tran_reward}
    \end{subfigure}
    \hfill
    \begin{subfigure}[t]{0.48\linewidth}
        \centering
        \includegraphics[width=\linewidth]{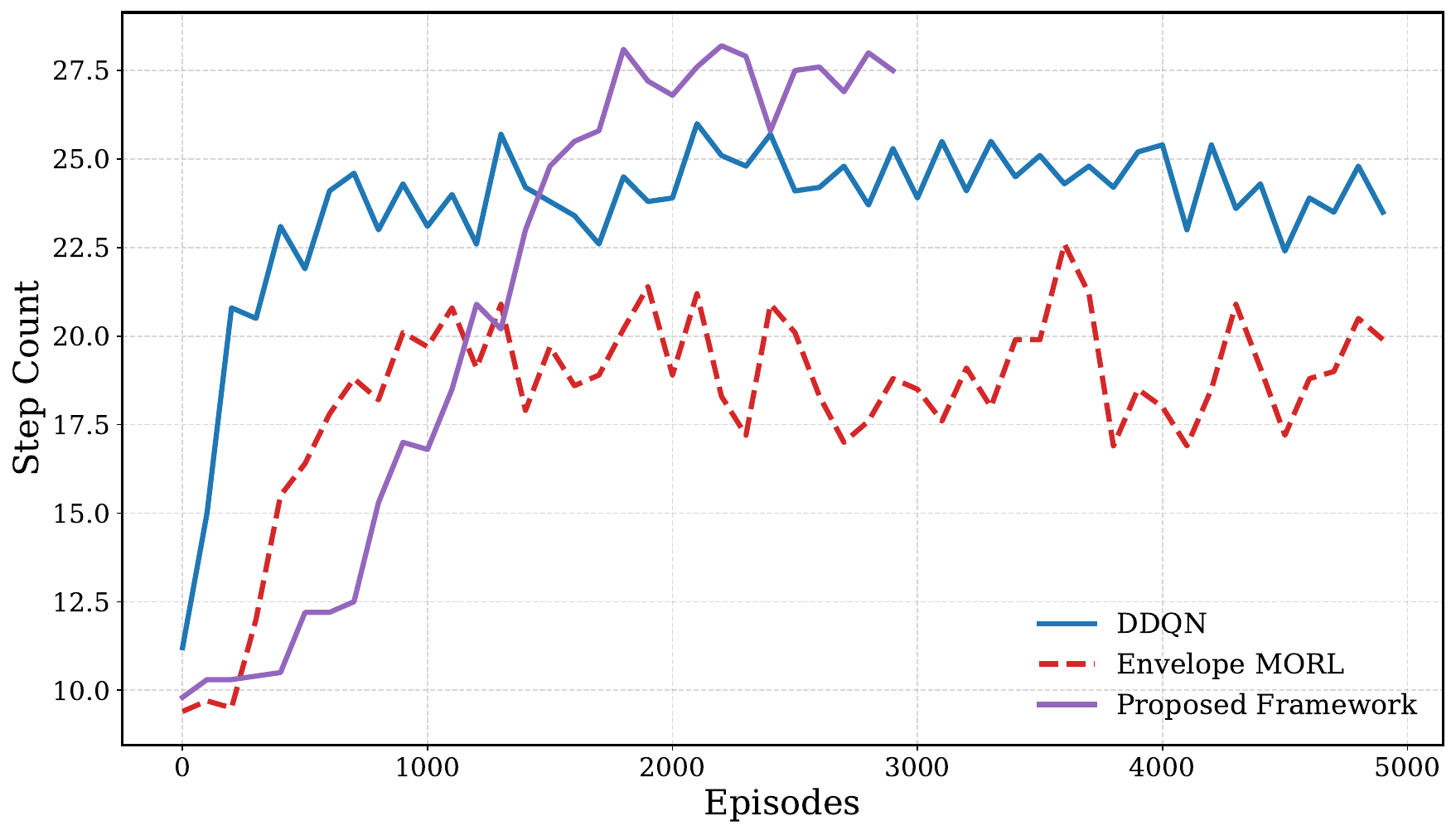}
        \caption{Survival (Episode Length)}
        \label{fig:step_count}
    \end{subfigure}

    \vspace{0.5em}

    \begin{subfigure}[t]{0.48\linewidth}
        \centering
        \includegraphics[width=\linewidth]{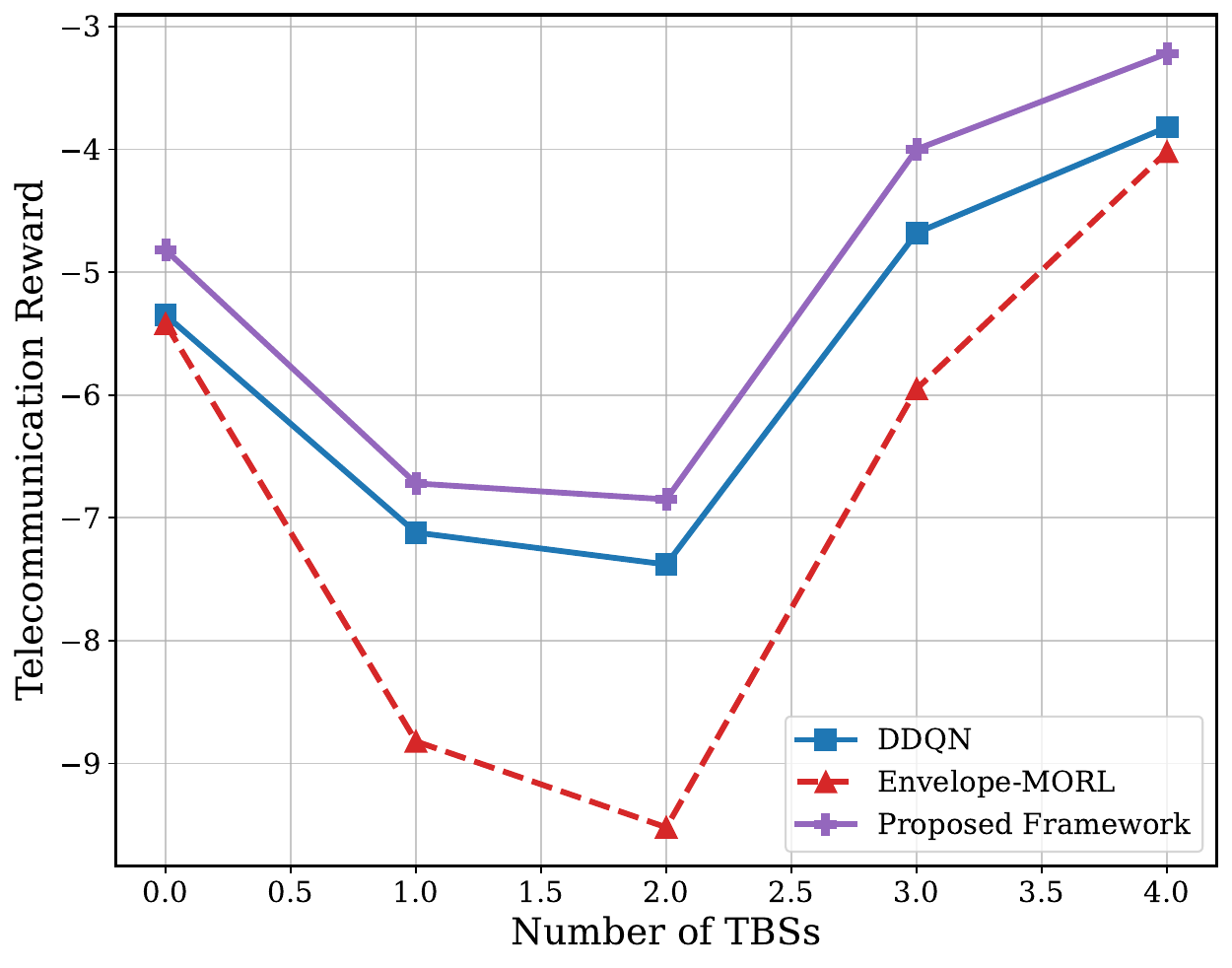}
        \caption{Communication Reward}
        \label{fig:tele_cost}
    \end{subfigure}
    \hfill
    \begin{subfigure}[t]{0.48\linewidth}
        \centering
        \includegraphics[width=\linewidth]{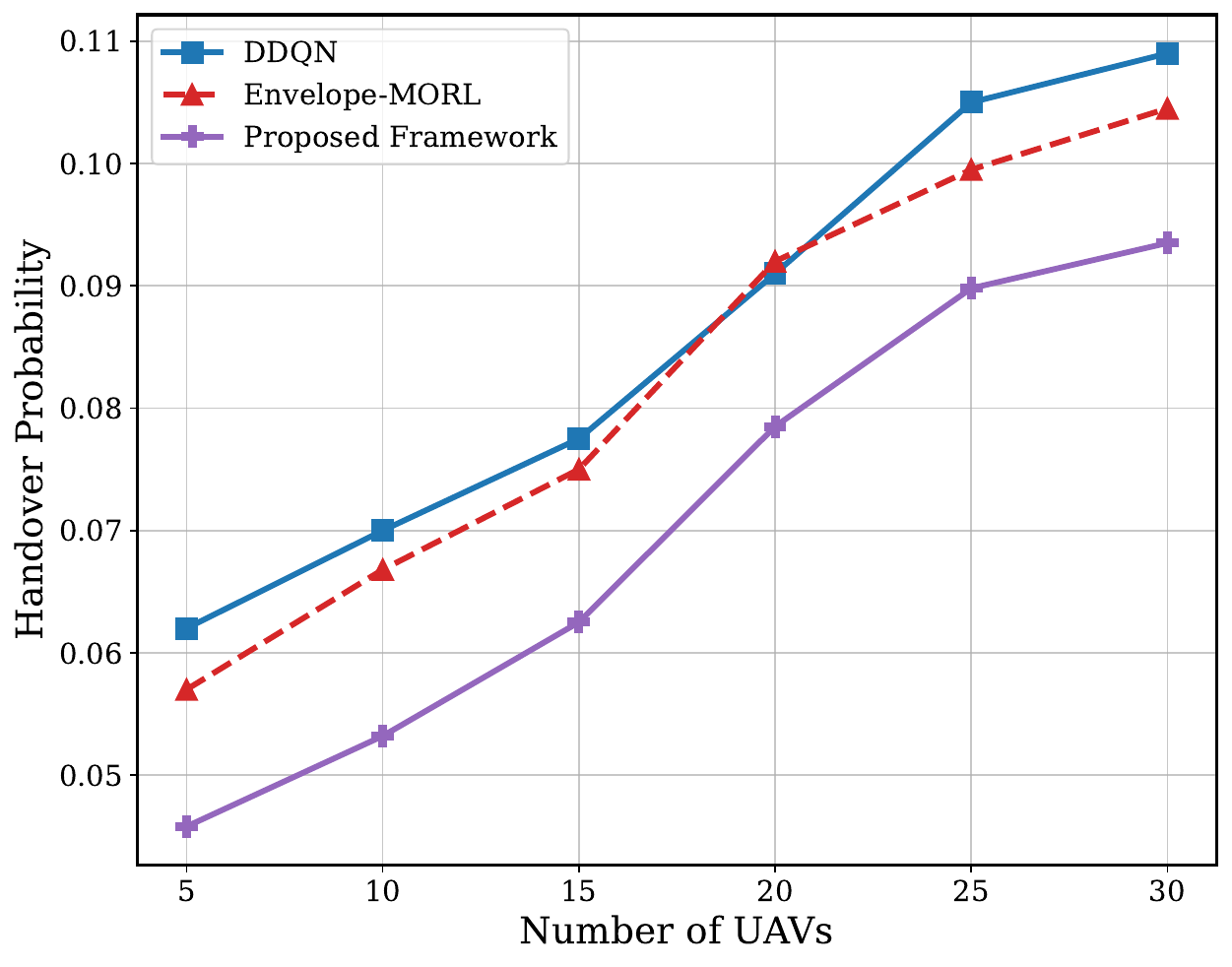}
        \caption{HO Probability}
        \label{fig:hos}
    \end{subfigure}

    \caption{Training convergence and scalability evaluation of the proposed Cloud-Edge Agentic framework.}
    \label{fig:combined_2x2}
\end{figure*}

\subsection{Simulation Results}

To evaluate the robustness in congested aerial corridors, Fig.~\ref{fig:combined_2x2} reports key metrics, i.e., transportation reward, survival time, communication reward, and HO probability. 


{
Fig.~\ref{fig:tran_reward} and Fig.~\ref{fig:step_count} illustrate the convergence of the transportation reward and the survival step count, respectively. As shown in Fig.~\ref{fig:tran_reward}, the conventional DDQN baseline learns quickly but saturates at a suboptimal local minimum. This is because standard DRL struggles with the large and sparse exploration space of joint 3D kinematics and network selection. In contrast, the proposed framework exhibits a steeper, more sustained learning curve, overtaking DDQN near episode 1,500 and achieving the highest steady-state reward. 

This superiority is directly tied to physical safety, as reflected by the survival step count in Fig.~\ref{fig:step_count}. Specifically, in DDQN, severe collisions cause early termination. Because the latter relies only on trial-and-error, it suffers from frequent early collisions. Conversely, when a UAV in our framework detects a high-risk state, the onboard edge-LLM immediately shifts the scalarization weights via cognitive reflection to prioritize the collision penalty ($\rho_{\text{crash}}$). This zero-shot guidance avoids early terminations and allows the proposed framework to converge to near-maximum episode lengths.

Fig.~\ref{fig:tele_cost} evaluates the communication reward as a function of the number of TBSs. Initially, without any TBS, UAVs are connected to the HAPS without any attenuation in the link quality. However, introducing a single or 2 TBSs creates severe interference and uncoordinated handover opportunities between the TBSs and HAPS, causing a sharp drop in the communication reward for all methods. However, as more TBSs are added to provide continuous coverage, the reward recovers and better coordination is achieved. The proposed framework consistently maintains the highest communication reward since the HAPS meta-controller intelligently coordinates associations to minimize cross-tier interference.

Fig.~\ref{fig:hos} illustrates the handover probability as the UAV swarm density increases from $M=5$ to $30$. As the airspace becomes congested, baseline DRL agents greedily chase the strongest instantaneous signal, resulting in erratic, high-frequency handovers. In contrast, the HAPS meta-controller anticipates congestion and issues zero-shot spatial offloading directives. This strategic coordination successfully suppresses unnecessary network switching, granting the proposed framework the lowest and most stable handover probability across all traffic densities.
}

\revised{


}

\section{Conclusion}
\label{sec:conclusion}
In this paper, we proposed a Cloud-Edge Agentic AI framework for jointly optimizing 3D UAV mobility and handover management in ITNTNs. By decoupling strategic reasoning and fast physical control across multiple timescales, the framework bridges LLM reasoning with real-time DRL execution. A HAPS-based meta-controller mitigates network congestion, while edge-LLMs improve local safety-aware decision-making. Simulation results show clear gains over conventional DRL in safety, learning efficiency, and system throughput. Future work will investigate lightweight quantization to reduce edge inference latency.
\bibliographystyle{IEEEtran}
\bibliography{main-clean.bib}
\end{document}